\documentclass{article}

\usepackage{arxiv}

\usepackage[utf8]{inputenc} 
\usepackage[T1]{fontenc}    
\usepackage{hyperref}       
\usepackage{url}            
\usepackage{booktabs}       
\usepackage{amsfonts}       
\usepackage{nicefrac}       
\usepackage{microtype}      
\usepackage{lipsum}		
\usepackage{graphicx}
\usepackage{natbib}
\usepackage{doi}

\usepackage{caption} 
\usepackage{subcaption}
\usepackage{multirow}
\usepackage{isomath}

\usepackage{amsmath, amsthm, amssymb}

\usepackage[font=normalsize,labelfont={bf}]{caption}
\usepackage{algpseudocode}
\usepackage{algorithm}

\title{Proprioceptive Learning with \\Soft Polyhedral Networks\thanks{A preprint submitted to the \textbf{International Journal of Robotics Research} for review.}}

\author{
    {Xiaobo Liu, Xudong Han}\\
    Department of Mechanical and Energy Engineering\\
    Southern University of Science and Technology\\
    Shenzhen, China 518055\\
    \And
    {Wei Hong}\\
    Department of Mechanics and Aerospace Engineering\\
    Southern University of Science and Technology\\
    Shenzhen, China 518055\\
    \And
    {Fang Wan*}\\
    School of Design\\
    Southern University of Science and Technology\\
    Shenzhen, China 518055\\
    \texttt{wanf@sustech.edu.cn} \\
    \And
    {Chaoyang Song*}\\
    Department of Mechanical and Energy Engineering\\
    Southern University of Science and Technology\\
    Shenzhen, China 518055\\
    \texttt{songcy@ieee.org} \\
}

\renewcommand{\shorttitle}{Proprioceptive Learning with Soft Polyhedral Networks}

\hypersetup{
pdftitle={Proprioceptive Learning with Soft Polyhedral Networks},
pdfsubject={q-bio.NC, q-bio.QM},
pdfauthor={Xiaobo Liu, Xudong Han, Wei Hong, Fang Wan, Chaoyang Song},
pdfkeywords={Soft Robotics, Force and Tactile Sensing, In-finger Vision, Proprioception},
}

\begin{document}
\maketitle

\begin{abstract}
    Proprioception is the ``sixth sense'' that detects limb postures with motor neurons. It requires a natural integration between the musculoskeletal systems and sensory receptors, which is challenging among modern robots that aim for lightweight, adaptive, and sensitive designs at a low cost. Here, we present the Soft Polyhedral Network with an embedded vision for physical interactions, capable of adaptive kinesthesia and viscoelastic proprioception by learning kinetic features. This design enables passive adaptations to omni-directional interactions, visually captured by a miniature high-speed motion tracking system embedded inside for proprioceptive learning. The results show that the soft network can infer real-time 6D forces and torques with accuracies of 0.25/0.24/0.35 N and 0.025/0.034/0.006 Nm in dynamic interactions. We also incorporate viscoelasticity in proprioception during static adaptation by adding a creep and relaxation modifier to refine the predicted results. The proposed soft network combines simplicity in design, omni-adaptation, and proprioceptive sensing with high accuracy, making it a versatile solution for robotics at a low cost with more than 1 million use cycles for tasks such as sensitive and competitive grasping, and touch-based geometry reconstruction. This study offers new insights into vision-based proprioception for soft robots in adaptive grasping, soft manipulation, and human-robot interaction.
\end{abstract}

\keywords{Soft Robotics \and Force and Tactile Sensing \and In-finger Vision \and Proprioception}

\section{Introduction}
\label{sec:Intro}

    Human fingers are dexterously adaptive in handling physical interactions through the bodily neuromuscular sense of proprioception expressed in multiple modalities. The neurological mechanism of proprioception is to sense from within, involving a complex of receptors for position and movement, as well as force and effort \citep{Taylor2009Proprioception}. Although rich literature has been devoted to the research of artificial skins \citep{You2020Artificial, Li2022Multifunctional, Wang2021Artificial} and robotic end-effectors \citep{Lee2020Twister, Odhner2014Compliant, Zhang2022Finger, Sun2021Artificial}, design integration of the two into a coherent robot system remains a challenge. The mechanical properties of human skin affect the activation of receptive organs, among which viscoelasticity is one of the most critical factors that are difficult to model \citep{Joodaki2018Skin}, resulting in time-dependent nonlinear behaviors \citep{Parvini2022Viscoelastic, Malhotra2019Linear, Wang2007InVivo}. With a growing trend in building soft robotic systems, designing soft fingers with proprioception extends the robot's adaptive intelligence while interacting with the physical world or human operators.

    We present the Soft Polyhedral Networks capable of vision-based proprioception with passive adaptations in omni-directions, significantly extending our previous work on vision-based tactile sensing with the soft robotic network \citep{Wan2022Visual}. 
    \begin{table}[t!]
        \small\sf\centering
        \caption{\textbf{A comparison between state-of-the-art soft sensory fingers and fingertips and our design.}}
        \label{tab:TactileSOTA}
        \begin{tabular}{lllcc}
            \hline
            \textbf{Sensor} &
              \textbf{Sensing Method} &
              \textbf{Geometric Adaptation} &
              \textbf{\begin{tabular}[c]{@{}c@{}}Compression \\ Force Range\end{tabular}} &
              \textbf{\begin{tabular}[c]{@{}c@{}}Precision in\\ Relative Error\end{tabular}} \\ \hline
            \begin{tabular}[c]{@{}l@{}}GelSight \\ \citep{Yuan2017Gelsight}\end{tabular} &
              Internal Vision &
              Regional at the Fingertip &
              $\sim$25.0 N &
              $\sim$2.7\% \\ \hline
            \begin{tabular}[c]{@{}l@{}}Insight\\ \citep{Sun2022Soft}\end{tabular} &
              Internal Vision &
              Regional as a Finger &
              2.0 N &
              $\sim$1.5\% \\ \hline
            \begin{tabular}[c]{@{}l@{}}Soft Continuum Finger\\ \citep{Thuruthel2019Soft}\end{tabular} &
              Strain Sensor &
              Global as a Finger &
              $\sim$0.35 N &
              15.3\% \\ \hline
            \begin{tabular}[c]{@{}l@{}}Fin Ray Finger (FRE) Model\\ \citep{Shan2020Modeling}\end{tabular} &
              Theoretical Model Only &
              Global as a Finger &
              $\sim$50.0 N &
              $\sim$13.7\%* \\ \hline
            \begin{tabular}[c]{@{}l@{}}FRE w/ External Vision\\ \citep{Xu2021Compliant}\end{tabular} &
              External Vision &
              Global as a Finger &
              3.0 N &
              6.5\% \\ \hline
            \begin{tabular}[c]{@{}l@{}}\textbf{Soft Polyhedral Network}\\ \textbf{(Ours)}\end{tabular} &
              Internal Vision &
              Global as a Finger &
              20.0 N &
              \textbf{1.25\%} \\ \hline
            \multicolumn{5}{l}{*Precision of the theoretical model proposed in \citep{Shan2020Modeling}, not a precision for sensing.}
        \end{tabular}
    \end{table}
    The design method transforms any polyhedral geometry into a soft network with mechanically programmable adaptation under passive interaction. In this study, we choose a particular design variation for robotic finger integration. By adding a miniature motion caption system to the base, we accurately captured and encoded the soft network's whole-body deformation in real-time by tracking the spatial movement of a fiducial marker attached inside. This allows us to quantitatively study the viscoelasticity in the soft metamaterial, which is usually ignored in soft robotics. To model the non-negligible viscoelasticity of the soft network for dynamic proprioception, we encode both deformation and kinetic input features to learn a more accurate data-driven model, which is not yet reported in other vision-based soft force sensors to the best of our knowledge, achieving state-of-the-art force sensing as shown in Table \ref{tab:TactileSOTA}. One can attach the proposed soft networks to almost any rigid grippers, or even soft ones, of compatible sizes to enable high-performing proprioception and omni-directional adaptation simultaneously at a low cost, accomplishing tasks such as sensitive and robust grasping against rigid grippers, impact absorption, and touch-based geometry reconstruction. The contributions of this work are listed as the following:
    \begin{itemize}
        \item Proposed a generic design method for a class of soft polyhedral networks with an embedded vision for proprioception.
        \item Implemented Sim2Real proprioceptive learning for adaptive kinesthesia to reproduce real-time physical interactions in 3D.
        \item Proposed visual force learning for viscoelastic proprioception with state-of-the-art 6D force (0.25/0.24/0.35 N) and torque (0.025/0.034/0.006 Nm) sensing.
        \item Demonstrated competitive capabilities of proprioceptive learning for achieving various fine-motor skills in object handling with robots even after 1 million use cycles.
    \end{itemize}

    The rest of this paper is organized as the following. Section \ref{sec:Method-AdaptiveDesign} introduces the proposed design of the soft finger with omni-directional adaptation and vision-based integration for sensing. Section \ref{sec:Method-Sim2Real} presents the method of integrating Finite Element Analysis and machine learning to reconstruct the finger's adaptive kinesthesia in real time for Sim2Real transfer. Section \ref{sec:Method-ViscoLearning} introduces the visual force learning method that leverages the material's viscoelasticity for static interaction and kinetic motion for dynamic grasping. Section \ref{sec:Results} presents experimental results that demonstrate the use of proprioceptive learning for impact absorption and touch-based geometry reconstruction. The conclusion, limitations, and future work are in the final section.

\section{Related work}
\label{sec:Related}

\subsection{Rigid and Soft Finger Adaptation}

    For industrial scenarios with task-specific needs, robotic fingers or grippers are usually fully actuated with just one or just a few degree-of-freedoms (DOFs) with rigid-bodied links or components. Inspired by human fingers, under-actuation and the integration of softness are widely appreciated when designing robotic fingers that are adaptive to the changes in object geometry \citep{Shimoga1996SoftI, Shimoga1996SoftII}, where the modeling of contact mechanics and friction limit surfaces enables one to study further the grasping and manipulation problems in robotics \citep{Xydas1999Modeling}. Previous work by \cite{Hussain2020Design} introduced the design method for a tendon-driven, under-actuated gripper with interpenetrating phase composite materials as flexible joints to achieve enhanced adaptation in grasping. Recent development in soft robotics promotes robotic fingers with a full-body soft design that conforms to the object geometry through fluidic actuation and passive adaptation. Recent work by \cite{Teeple2020Multi} presented a soft robotic finger with multi-segmented actuation for enhanced adaptation and dexterity in object manipulation. Further introduction of adaptiveness in the robotic palm is investigated by \cite{Subramaniam2020Design}, where the coupling effects of a soft robotic palm further enhance grasping robustness. Discussion on the softness distribution index by \cite{Naselli2021SoftNess} provides a working guideline for designing and modeling soft-bodied robots that are generally applicable to soft continuum manipulators and soft fingers. 

    The Fin Ray Effect (FRE) soft finger is a design with an excellent adaptation that effectively transforms any industrial gripper into a soft robotic hand. The readers are encouraged to refer to the work by \cite{Shan2020Modeling} for an in-depth review of the related literature and its theoretical modeling. While the FRE finger can provide geometric adaptation in the 2D plane for grasping, recent work shows a novel finger network design capable of omni-directional adaptation in 3D \citep{Yang2020Rigid}, shown as the Pyramid variations in Figure \ref{fig1}A. While equally adaptive to geometric adaptation on the primary interaction face, the finger network design can also produce geometric adaptation from the sideways or on the edge for adaptive grasping \citep{Wan2020AReconfigurable}. It provides a generic design method for a wide range of soft robotic finger networks with similar adaptation in omni-directional interactions \citep{Song2022Robotic}. And it also features a hollow volume inside to visually capture its geometric deformation process, which can be integrated with either optical fibers \citep{Yang2021Learning} or miniature cameras \citep{Wan2022Visual} for accurate tactile sensing integration. Compared to the FRE finger, the omni-directional adaptation behavior of the omni-finger becomes much more challenging to solve mechanically through analytical modeling, where a data-driven method integrating machine learning \citep{Tapia2020MakeSense} and finite element analysis \citep{Duriez2013Control, Largilliere2015RealTime} could be a potential solution.

\subsection{Sensory Integration during Soft Contact}

    Scientific literature reports a wide range of sensory integration in robotic manipulation by estimating the soft material's passive adaptation during contact, including 1) soft fingertips with surface adaptation in the local regions \citep{Lambeta2020Digit, Shimoga1996SoftI, Shimoga1996SoftII} and 2) soft fingers with structural adaptation in the global spaces \citep{Truby2018Soft, Subramaniam2020Design, Teeple2020Multi}. While both are specifically designed for robotic applications, the artificial skin represents another research stream aiming at a broader range of applications for human-machine interactions \citep{Yan2021Soft, Zhu2021Skin}. The \textit{soft robotic fingertip} is widely adopted to capture localized surface deformations during contact. Sensors with one or multiple modalities can be embedded under a small piece of soft material and molded to the size of a fingertip \citep{Wettels2014Multimodal, Park2015Fingertip}. However, recent research shows a growing adoption of visual sensing by tracking the soft materials’ surface deformation \citep{Yamaguchi2016Combining, Yuan2017Gelsight}. This strategy significantly reduces design complexity and integration cost while generating a rich perception of contact \citep{Sun2022Soft}. It should be noted that many soft robotic fingertips are equivalent to artificial skins but with integrated designs packed in a small form factor for convenient installation at the end of existing grippers or fingers.

    The \textit{soft robotic finger} represents another approach that involves active or passive actuation of the soft body deformation on a global scale to replace the rigid gripper mechanism for grasping. One could directly integrate artificial skins into robot fingers for the same purpose \citep{Zhu2021Skin, Heo2020Human, Liu2022Star}. Many soft robotic fingers leverage both the soft materials' active and passive deformations and can integrate multiple modalities for tactile sensing \citep{Truby2018Soft, Kim2020Sustainable}. Under fluidic \citep{Terryn2017Self, Hu2020Bioinspired} or electrical \citep{Li2019Dielectric, Acome2018Hydraulically} actuation, the soft-bodied finger can actively generate geometric deformation to produce a grasping action. During contact, the soft robotic finger can passively conform to the object's geometry \citep{Cheng2022Centrifugal, Liu2020Two}. Recent work by \cite{Wall2023Passive} shows a sensorization method for soft pneumatic actuators that uses an embedded microphone and speaker to measure different actuator properties. Machine learning algorithms may also be applied to estimate soft body deformations \citep{Hu2023Stretchable, Loo2022Robust, Scharff2021Sensing}. In summary, there remains a challenge in achieving simultaneous contact perception and globalized grasp adaptation at a reduced cost and design complexity for fine motor controls.

    Many soft robots are made from polymers such as plastics, rubber, and silica gel \citep{Hu2023Stretchable, Cecchini20234D} or metamaterials with structural compliances \citep{Xu2019Optical, Wan2022Visual}. Analytical methods using pseudo-rigid-body models (PRBMs) are inherently limited in mechanic assumption to predict the physical interactions accurately \cite{Shan2020Modeling}. Viscoelasticity characterizes a time-dependent deformation among soft robots, leading to stress relaxation and creep that are difficult to model \citep{Gutierrez-Lemini2013Engineering}. In applications where the soft sensor bears dynamic loadings, dynamic hysteresis affects its measurement accuracy \citep{Zou2018High, Oliveri2018Model}. Difficulties in representing and detecting soft materials' complex volumetric deformations make it challenging to study viscoelasticity. Many vision-based sensors use a layer of soft skin to isolate the camera from the environment, aiming for stable detection of the interaction physics, such as the intensity of reflective light and marker displacement \citep{Yuan2017Gelsight, She2020Exoskeleton}. These sensors mainly capture local deformations on the interaction surfaces. For sensors where the camera is open to the environment, the tracked motions are usually limited to planar movements, and the detection is subject to occlusions \citep{Xu2021Compliant}.
    
\section{Soft Polyhedral Networks with Embedded Vision for Proprioception}
\label{sec:Method-AdaptiveDesign}

\subsection{Soft Polyhedral Networks}

\begin{figure*}[!ht]
        \centering
        \includegraphics[width=\textwidth]{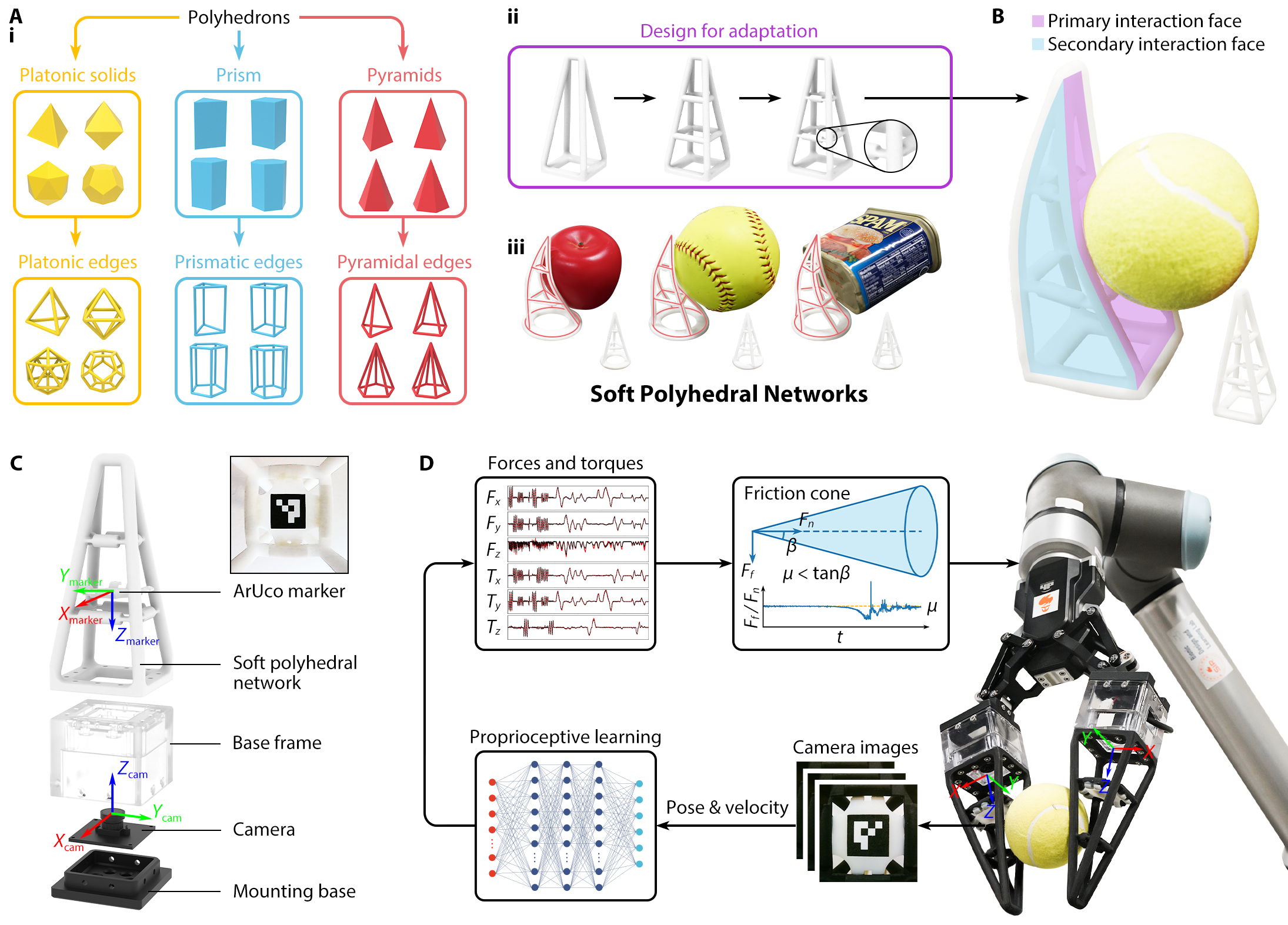}
        \caption{
        \textbf{Soft Polyhedral Network design with embedded vision.} 
        (A) A generic design process applicable to all polyhedrons, (i) starting with removing all faces and replacing all edges with beam structures made from soft materials, then (ii) adding layers inside with the flexure joints, resulting in (iii) a class of soft networks that are geometrically adaptive to external interactions. 
        (B) An enhanced version of the Soft Polyhedral Network with a primary interaction face (marked in pink) and a secondary interaction face (marked in blue). The primary face has an extended contact area with a trapezoid frame, and the secondary face enables adaption in 3D. 
        (C) Exploded view for vision integration by mounting the soft network on top of a base frame housing a high-speed miniature camera, capturing the soft network's 6D motion during adaptation by tracking an ArUco marker attached inside. 
        (D) The pipeline for proprioceptive learning when using the Soft Polyhedral Network as fingers of a common gripper system. The camera captures the spatial deformation of the soft network by tracking the ArUco marker's 6D movement. We feed pose and velocity inputs to a neural network to infer 6D forces and torques as the output, which can be further processed to estimate the gripping and shear forces and fed to the robot control loop for reactive object manipulation based on the friction cone model.}
        \label{fig1}
    \end{figure*}

    A polyhedron is generally understood as a solid geometry in three-dimensional space, featuring polygonal faces connected by straight edges, including prisms, pyramids, and platonic solids \citep{Demaine2007Geometric}. Inspired by recent development in soft robotics, we propose a generic design method by turning all edges of a polyhedron into beam structures made from soft materials, then adding layers inside to form a network, followed by redesigning the ends of all mid-layer edges as flexure joints to reduce inferences during deformation while providing sufficient structural support in a compliant manner in Figure \ref{fig1}A. 
    The resultant designs exhibit excellent adaptations in 3D, formulating a class of Soft Polyhedral Networks. This design method is generic, as one can reconfigure the parameters to fine-tune the soft network's passive adaptation.  In this study, we chose the pyramid shape as the base design and modified it with two vertices on top. Figure \ref{fig1}B shows this design features a primary interaction face for typical grasping and a secondary one to enable spatial adaptation, such as 3D twisting. The resultant structure exhibits a large, hollow volume inside with an unobstructed view from the bottom, allowing a direct capture of the adaptive deformations during physical interaction. To attain stable and homogeneous performance, we fabricated the whole network through vacuum molding using polyurethane elastomers (Hei-cast 8400 from H\&K) with a mixing ratio of 1:1:0 for its three components to achieve 90A hardness. Alternatively, one can turn to direct 3D printing with TPU or other compliant materials for fabrication \citep{Yi2018Customizable}.

\subsection{Embedded Motion Tracking}
\label{sec:Methods-DesignAndFab}
    We embedded a miniature motion-tracking system inside the Soft Polyhedral Network to mimic a proprioceptor. As shown in Figure \ref{fig1}C, the system involves a high-speed camera of up to 330 fps (manually adjustable lens, Chengyue WX605 from Weixinshijie) with a large viewing angle (170°) fixed on a mounting base inside the network and a plate attached to the network's first layer with a fiducial marker (ArUco of 16 mm width) stuck to its bottom. The soft network's spatial adaptation is expressed by its structural compliance, then filtered by the fiducial marker's spatial movement inside, next captured by the high-speed camera as image features, and finally encoded as a time series of dimensionally reduced 6D pose vector $\mathbf{D}_t = (D_x, D_y, D_z, D_{rx}, D_{ry}, D_{rz})_t$, namely, the translation and rotation of the marker relative to its initial pose $p_0$ before any deformation. The motion tracking system has a high precision of up to 0.005 mm and 0.018$^\circ$. Table \ref{tab:TrackingStability} shows that the marker detection has excellent stability at all three resolutions. The system achieves a 100$\%$ success detection rate in a test of 8,000 consecutive frames, even at the lowest resolution. In the rest of the paper, we always set the resolution to 640$\times$360 at 330 fps.
    \begin{table}[!ht]
        \small\sf\centering
        \caption{\textbf{Tracking stability of the embedded miniature motion capture system.}}
        \label{tab:TrackingStability}
        \begin{tabular}{lcccc}
            \toprule
            \multicolumn{1}{c}{\multirow{2}{*}{\begin{tabular}[c]{@{}c@{}}\textbf{Camera} \\ \textbf{Resolution}\end{tabular}}} &
              \multirow{2}{*}{\begin{tabular}[c]{@{}c@{}}\textbf{Distance} \\ \textbf{(mm)}\end{tabular}} &
              \multirow{2}{*}{\begin{tabular}[c]{@{}c@{}}\textbf{Computation Time} \\ \textbf{(ms)}\end{tabular}} &
              \multicolumn{2}{c}{\textbf{Range of Fluctuations}} \\ 
            \multicolumn{1}{c}{} &      &      & \textbf{Positional Noise (mm)}     & \textbf{Angular Noise (degree)}    \\ 
            \midrule
            640$\times$360       & 45.6 & 3.7  & $[$0.014, 0.017, 0.022$]$ & $[$0.573, 0.514, 0.080$]$ \\
            1280$\times$720      & 45.2 & 10.1 & $[$0.015, 0.011, 0.019$]$ & $[$0.428, 0.214, 0.041$]$ \\
            1920$\times$1080     & 44.7 & 16.4 & $[$0.005, 0.009, 0.020$]$ & $[$0.242, 0.115, 0.018$]$ \\ 
            \bottomrule
        \end{tabular}
    \end{table}
    
    We adopt the motion capture solution for its simplicity, transferability, and low cost in mechanical design and algorithmic computation. For example, the motion capture solution can be easily transferred to Soft Polyhedral Networks other than the pyramid shapes, including the prism and platonic ones. One can easily mount the Soft Polyhedral Networks on standard grippers by replacing its current rigid fingertips (Figure \ref{fig1}D). The system proposed in this study involves only three components: the Soft Polyhedral Network, a miniature high-speed camera, and a pair of base frame and mounting base for fixturing. Simplicity in design is the enabling factor of the proposed Soft Polyhedral Network, supporting its robust adaptation with vision-based tactile sensing for robotic manipulation.   

\section{Learning Adaptive Kinesthesia}
\label{sec:Method-Sim2Real}

\subsection{Stiffness Distribution and FEM Simulation}
\label{sec:Methods-MeasureStiffness}
    
    We conducted a series of unidirectional compression experiments to estimate the stiffness distribution of the Soft Polyhedral Network defined as force over displacement.
    \begin{figure}[!ht]
        \centering
        \includegraphics[width=0.6\columnwidth]{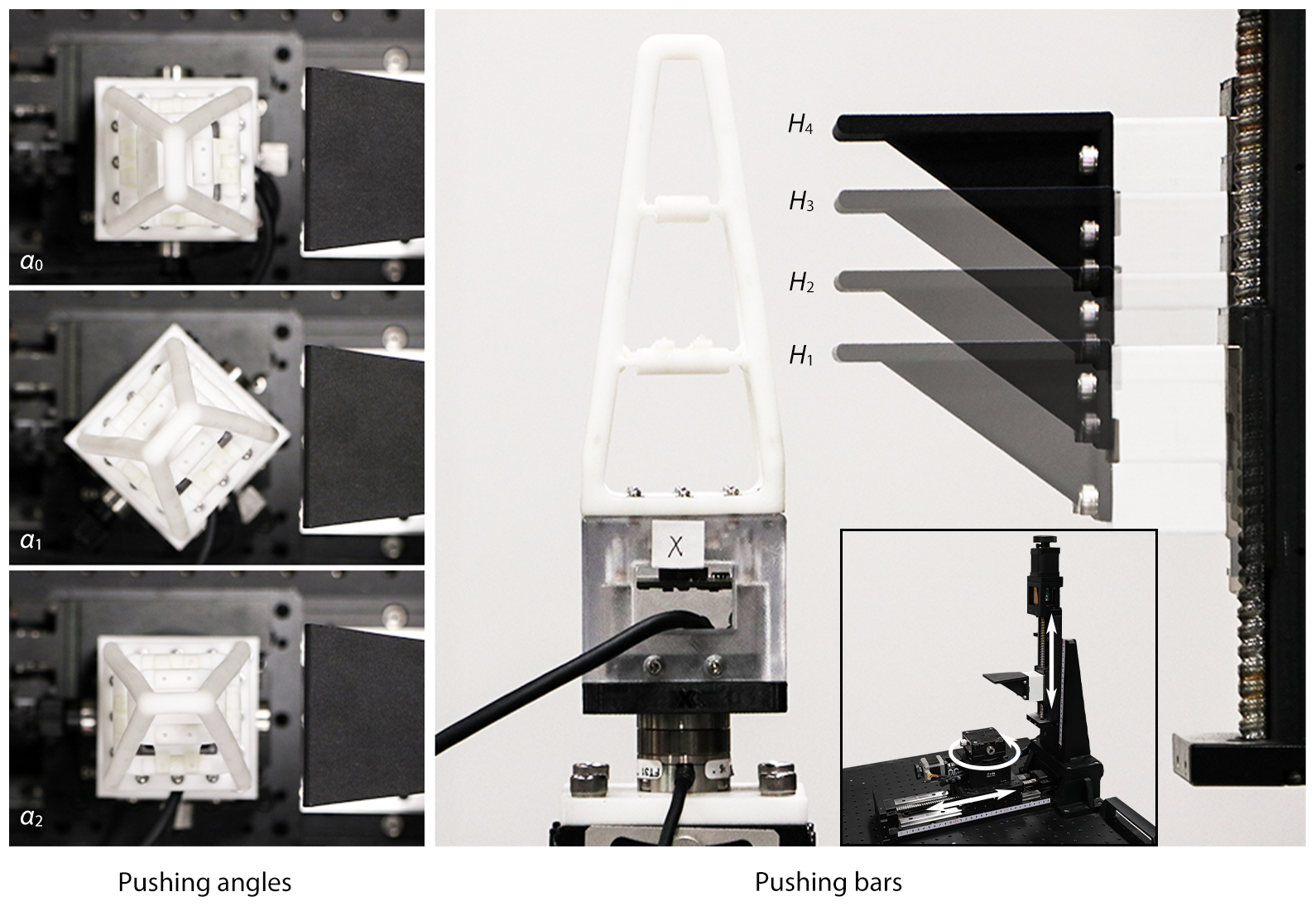}
        \caption{
        \textbf{Experiment setup for measuring stiffness.} 
        }
        \label{fig2}
    \end{figure}
    Figure \ref{fig2} shows that the soft finger is mounted on a high-performance force/torque sensor (Nano25 from ATI) on top of a custom test rig with two motorized linear motions and two manually driven rotary motions. The force/torque sensor has a resolution of 1/48 N for $F_x/F_y$, 1/16 N for $F_z$, 0.76 Nmm for $T_x/T_y$, and 0.38 Nmm for $T_z$. The probe compressed the Soft Polyhedral Network horizontally at 3 mm/s to a pre-defined depth of 15 mm. We conducted the experiments at three different pushing angles ($\alpha_0$, $\alpha_1$, and $\alpha_2$), where $\alpha_0 = 0^\circ$ is to compress the primary interaction face, $\alpha_1 = 45^\circ$ is to compress the edge between the primary and secondary interaction faces, and $\alpha_2 = 90^\circ$ is to compress the secondary interaction face. Meanwhile, we also adjust the compression height between $H_1$ to $H_4$. The push displacement $\delta$ and reaction force $F$ were recorded to calculate the corresponding stiffness $k = F/\delta$. 

    \begin{figure}[b!]
        \centering
        \includegraphics[width=0.6\columnwidth]{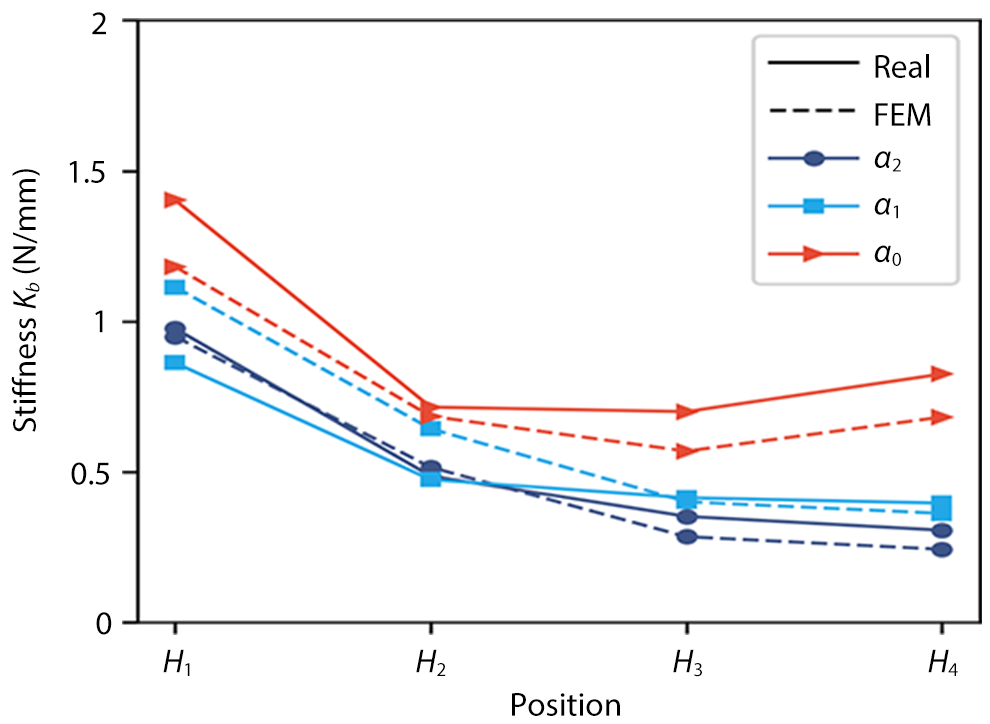}
        \caption{
        \textbf{Stiffness distribution of the Soft Polyhedral Network measured using the test rig and FEM.} 
        }
        \label{fig3}
    \end{figure}
    
    We used linear elastic elements and non-linear geometry in FEM simulations to model the large adaptive deformation. Calibrated to match the experimental stiffness measurement, the FEM simulation used Young's modulus of 12.05 MPa, Poisson ratio of 0.5, and density of 11.3 $\rm g/cm^3$, and the solid elements in FEM are 10-node quadratic tetrahedrons with hybrid formulation (C3D10H). The plate for the fiducial marker is a much more rigid body with Young's modulus of 2,600 MPa. The Soft Polyhedral Network's bottom is fixed. A total of about 13,000 elements were used in the simulation. The stiffness distribution calculated with simulated data agrees well with the actual measurement in Figure \ref{fig3}, demonstrating a good match between simulation and the soft physical networks. Both measurements share a similar trend where a U-shape stiffness distribution suggests that the primary interaction face is highly adaptive with a conforming geometry during physical interaction. A decreasing stiffness distribution suggests that the edge and secondary interaction face are moderately adaptive. The average absolute error is 0.098 N/mm, and the average relative error is 15.12\%. For all experiments, Figure \ref{fig3} shows a decreasing stiffness distribution along the $z$-axis at $\alpha_1$ and $\alpha_2$. At $\alpha_0$, however, the stiffness decreases from 1.4 N/mm at $H_1$ to a minimum of 0.7 N/mm at $H_3$. and increases slightly to 0.825 N/mm at $H_4$. This unique stiffness distribution differs from that of the fin ray effect finger, where the stiffness at the fingertip drops to about 25\% of the stiffness near the base \citep{Shan2020Modeling}. 
    
    \begin{figure}[!ht]
        \includegraphics[width=\textwidth]{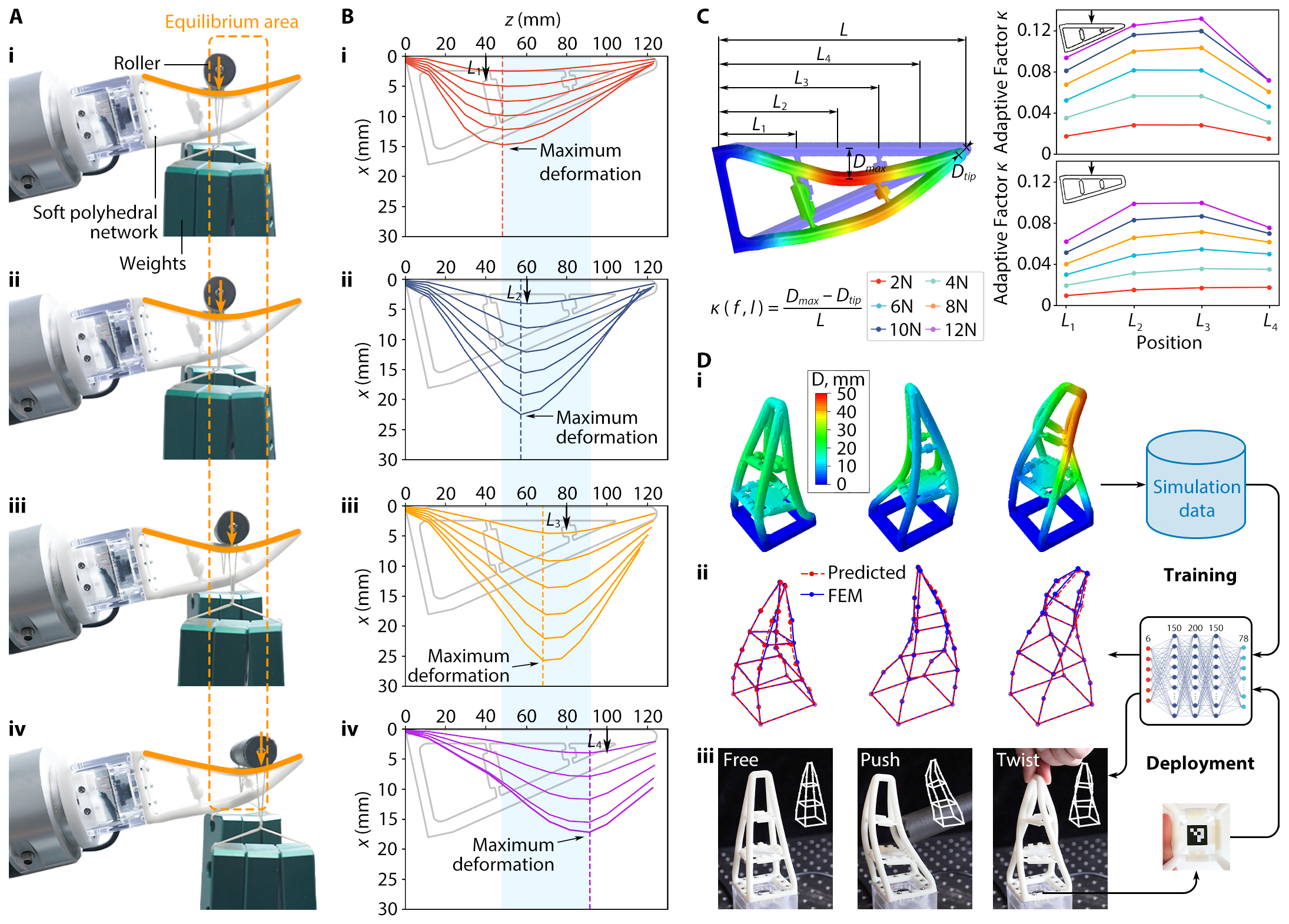}
        \caption{
        \textbf{Learning adaptive kinesthesia with Sim2Real proprioception.} 
        (A) Test for the soft network's passive adaptation by placing a roller at four different locations (i)$\sim$(iv) on the primary interaction face. The roller moves towards an equilibrium area marked in orange dashed lines. 
        (B) FEM simulations of the primary interaction face's adaptive deformation when applying 2$\sim$12 N forces at four initial contact locations marked with arrows. Note that the maximum deformation always occurs within an equilibrium region marked in light blue. 
        (C) Measurement of the adaptive factor $\kappa$ for the primary and secondary interaction faces. Both faces exhibit passive adaptation with $\kappa$ maximizing near $L_3$, resulting in an enclosed adaptation of the soft network upon external compression. Note that the adaptive capability in the primary interaction face is greater than the secondary one. 
        (D) After (i) collecting FEM simulation data of the soft network under external compressions at various angles and magnitudes, (ii) we train a Sim2Real multi-layer perceptron (MLP) to reproduce the spatial movement of 26 key points on the soft network. (iii) When deployed to the soft network prototype, the MLP predictions align well with observations in free-standing, pushing, and twisting scenarios.}
        \label{fig4}
    \end{figure}

    We investigated the soft network's passive adaptation by placing a 3D-printed roller of 7 mm in diameter with different weights (380$\sim$1140 g) at various locations of the primary interaction face along the horizontal direction. Results in Figure \ref{fig4}A show that the roller, supported by a ball bearing on each end, always rolled toward an equilibrium area to a complete stop. During the process, the Soft Polyhedral Network started deforming at the point of contact with a tendency to enclose the roller. This tendency generates an equilibrium area with the highest bending curvature between $L_2$ and $L_4$ in Figure \ref{fig4}B, causing the roller to rotate towards the lowest point until an equilibrium state. We studied the same interaction processes by simulation using Finite Element Methods (FEM). Results show that the maximum passive adaptation of the Soft Polyhedral Network always occurs within the shaded area in Figure \ref{fig4}B along the primary interaction face. The resultant spatial compliance is mechanically adaptive to external loadings, which we call ``adaptive kinesthesia'' of the Soft Polyhedral Network. Here, we define an adaptive factor $\kappa$ to measure adaptive kinesthesia under an external force $f$ exerted at location $l$ along the primary interaction face $S_i$ as
    \begin{equation}
        \kappa_i(f, l)= \frac{D_{max}(l')-D_{tip}}{L},\label{eq1}
    \end{equation}
    where $D_{max}$ is the maximum displacement of the adaptive deformation, $l'$ is the location of maximum displacement, and $D_{tip}$ is the tip displacement. The adaptive factor $\kappa$ reflects how well the network encloses objects along the primary interaction face, with a higher value indicating a better adaption. Figure \ref{fig4}C shows the simulated profiles of adaption for both primary and secondary interaction faces using FEM. All curves share a similar shape, and the adaptive factor $\kappa$ maximizes near $L_3$. We also found that the primary interaction face has better adaptability than the secondary one with $\kappa_1$ being 1.70, 1.59, 1.48 times $\kappa_2$ at $L_1$ to $L_3$. The adaptive factor at $L_4$ was measured at a similar amount (0.94 times), indicating that the segment towards the tip of the network is not as adaptive as the middle but behaves more like a rigid fingernail, which is desirable to produce a firm grasp. The high stiffness at the tip leads to greater adaptive compliance for the pyramid design of the soft network.
    
\subsection{Sim2Real Proprioceptive Learning} 

\begin{figure}[!ht]
        \centering
        \includegraphics[width=0.6\columnwidth]{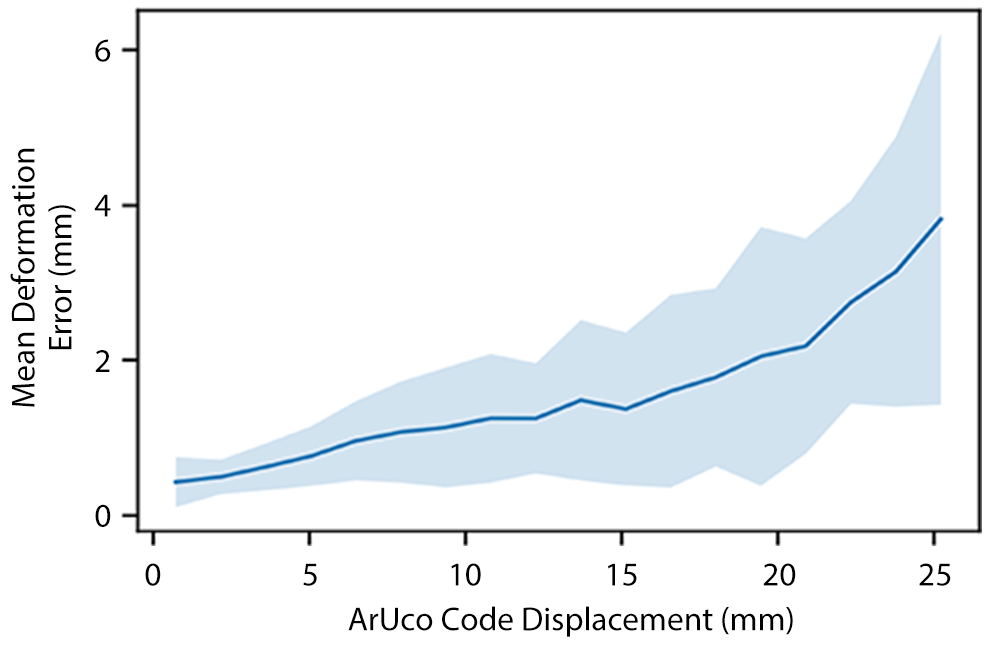}
        \caption{
        \textbf{Mean error distribution of Sim2Real learning for adaptive kinesthesia.} 
        The shaded area indicates one standard deviation of the average positional error.
        } 
        \label{fig5}
    \end{figure}

    Kinesthesia is appreciated as the ability to detect active or passive limb movements about a joint, which corresponds to the detection and reproduction of structural movement in the Soft Polyhedral Network during spatial interactions. We propose a Sim2Real learning strategy to detect and reproduce the Soft Polyhedral Network's adaptive kinesthesia, i.e., the passive proprioception of whole-body movement, using the embedded miniature camera for sensing and FEM data for training. As shown in Figure \ref{fig4}D(i), we collected training data from 12,000 simulations of a soft network model under various loading conditions. The geometry of the simulated soft network is represented by a collection of 26 feature points $\mathbf{M}=\{N_i:(x_i,y_i,z_i)|i = 1, …, 26\}$ as shown in Figure \ref{fig4}D(ii), including the intersections of all edges and the mid-points in between. We recorded the coordinates of these feature points and the fiducial marker's corresponding displacement $\mathbf{D}_t$. Assuming that the fiducial marker's spatial movement contains sufficient information to infer the soft network's adaptive deformation, we trained a regression model based on multi-layer perceptron (MLP) with $\mathbf{D}_t$ as input and the flattened $\mathbf{M}$ as output. The MLP has three hidden layers with 150, 200, and 150 neurons, respectively. The proprioceptive model is evaluated by the positional error $\sum_{i=1}^{26} \| \hat{N_i}-N_i\|/26$ where $\hat{N_i}$ is the predicted position of the simulated node $N_i$ as shown in Figure \ref{fig5}. 
    The average positional error grows as the soft network exhibits large-scale deformations during physical interactions, ranging from 0.4 to less than 4 mm, with an overall average of 1.18 mm. We applied the model trained from simulated data to a real soft network. Each prediction costs 0.4 ms on a laptop with NVIDIA GeForce GTX 1060. We made real-time predictions of its whole-body movement during physical interactions in Figure \ref{fig4}D(iii), demonstrating the power of Sim2Real learning of the soft network's proprioception in adaptive kinesthesia enhanced by machine learning with FEM (See supplementary material Movie S1).
     
\section{Sensing Force and Torque}
\label{sec:Method-ViscoLearning}

    \begin{figure*}[!h]
        \includegraphics[width=\textwidth]{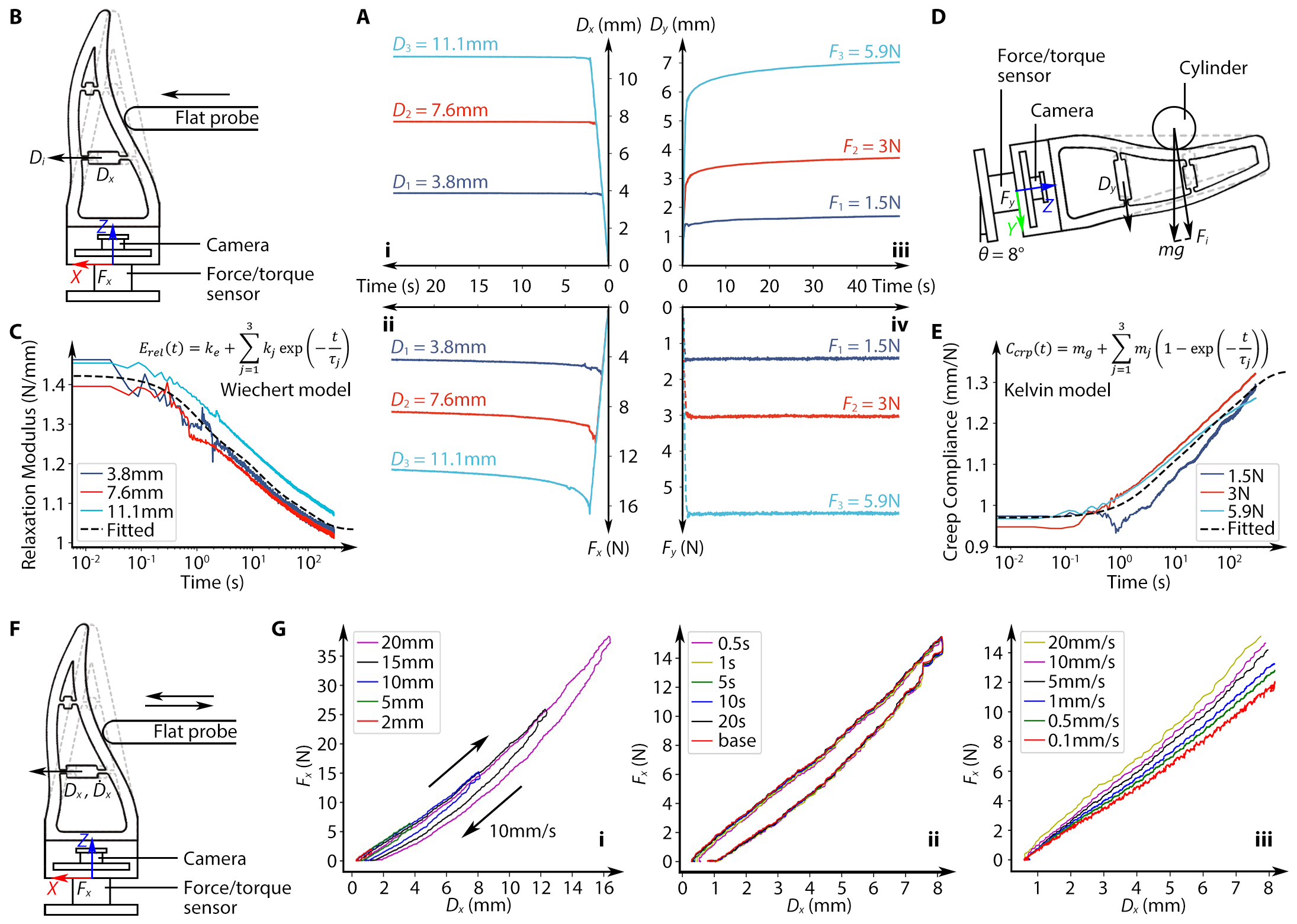}
        \caption{
        \textbf{Viscoelastic analysis of the Soft Polyhedral Network.} 
        (A) Results on the stress relaxation and creep experiments, including i) displacement and ii) force responses by a flat probe along the horizontal direction against the primary interaction face at three fixed distances for relaxation, as well as iii) displacement and iv) force responses by a cylindrical rod along the vertical direction against the secondary interaction face with three different weights for creep. (B) Experiment setup for stress relaxation with a Soft Polyhedral Network fixed vertically on top of a 6-axis FT sensor. (C) Measured relaxation modulus as a function of time and the fitted Wiechert model. (D) Experiment setup for creep test with a Soft Polyhedral Network fixed at $\gamma$ = 8° to keep the primary interaction face horizontal. (E) Measured creep compliance as a function of time and the fitted Kelvin model. (F) Experiment setup for dynamic loadings against the primary interaction face with a Soft Polyhedral Network fixed vertically on top of a 6-axis FT sensor. (G) Displacement and force responses with a flat probe compressing and de-compressing at different (i) loading depths, (ii) waiting times between cycles, and (iii) speeds.}
        \label{fig6}
    \end{figure*}

\subsection{Viscoelasticity Analysis in Static and Dynamic Interactions} 
\label{sec:Method-ViscoelasticInteractions}

    Viscoelasticity describes a material's characteristic to act like a solid and a fluid, which is universally applicable to robots made from soft matter. The metamaterial design and use of polyurethane for fabrication make the Soft Polyhedral Network responsive to adaptation in a time-dependent and rate-dependent manner. Meanwhile, recent literature suggested including the sense of velocity to extend the concept of proprioception \citep{Ager2020Proprioception}. In this section, we report results to characterize the Soft Polyhedral Network's viscoelastic behaviors in stress relaxation, creep, and dynamic loadings at various interaction speeds.

\begin{figure}[!ht]
        \centering
        \includegraphics[width=0.6\columnwidth]{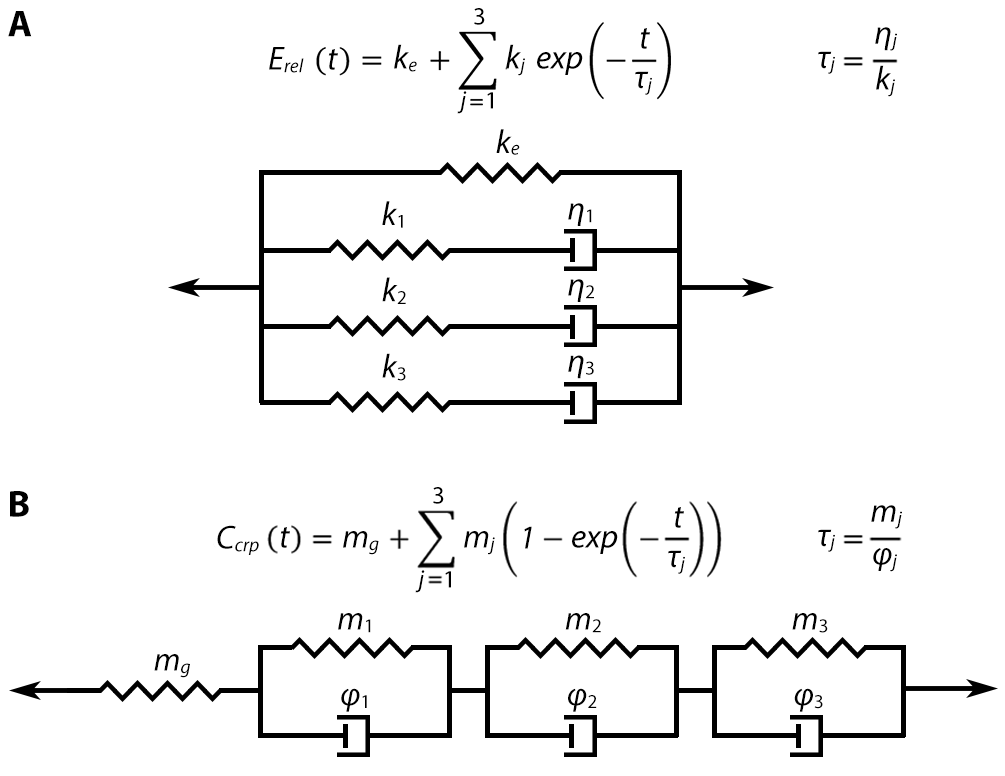}
        \caption{
        \textbf{The Wiechert model for stress relaxation in (A) and the Kelvin model for creep in (B).} 
        }
        \label{fig7}
    \end{figure}

    We started by investigating the Soft Polyhedral Network's stress relaxation to model its reaction force $\sigma(t)$ at a longer time scale of $t$ seconds by applying a constant strain $\epsilon_0$ at a room temperature of 25 $^{\circ}$C \citep{Gutierrez-Lemini2013Engineering}. We set up the experiment by mounting the soft network on a vibration isolation table with a 6-axis force-torque sensor (Nano25 from ATI) in between, as shown in Figure \ref{fig6}B. A 3D-printed, 5 mm thick, flat probe installed on the tool flange of a robot arm (UR10e from Universal Robots) was used to horizontally compress the soft network's primary interaction face at height $H_2$ to a certain depth $d$ and held the compression for 300 s. We recorded the marker pose and force/torque readings. Figure \ref{fig6}A(i) shows the soft network's $x$-axis displacement over time, where it immediately reaches a stable deformation as the compression completes. Simultaneously, the reaction force $F_x$ reaches the maximum in Figure \ref{fig6}A(ii), demonstrating the soft network's geometric adaptation during physical interactions. Then, $F_x$ starts to decrease exponentially until equilibrium. Figure \ref{fig6}C shows the relaxation modulus curves $E_{rel}(t) = \sigma(t)/\epsilon_0$ at three different depths. We model the stress relaxation process using a Wiechert model, composed of an elastic spring of stiffness $k_e$ parallel to three Maxwell elements as shown in Figure \ref{fig7}A. Each Maxwell element consists of a Hookean spring of stiffness $k$ and a Newtonian dashpot of viscosity $\eta$ connected in series, resulting in a characteristic time $\tau =\eta/k$. The fitted Wiechert model is described by
    \begin{equation}
        E_{rel}(t)= k_e + \sum_{j=1}^3 k_j \exp{(-t/\tau_j)}, \label{eq2}
    \end{equation}
    where the elastic modulus $k_e$ = 1.03 N/mm, the three Maxwell components $k_1$, $k_2$, $k_3$ = 0.15, 0.13, 0.11 N/mm , and their characteristic relaxation times $\tau_1$, $\tau_2$, $\tau_3$ = 1.0, 12.1, 109.5 s, respectively. 
    Such relaxation response characterizes the soft network's viscoelastic behavior, demonstrating adaptations at both geometric and molecular levels. The equilibrium modulus $E_{rel}(\infty)$ drops by 27$\%$ compared to the initial modulus $E_{rel}(0)$. This result indicates that for grasping tasks where the fingers must constantly hold the object, especially when the fingers are made from soft materials, the grasp planning algorithm should anticipate a diminishing gripping force due to viscoelastic relaxation to avoid dropping. Current solutions for object manipulation with soft robotic fingers usually adopt an open-loop control to overconstrain the object's movement with form closure \citep{Manti2015Bioinspired, Zhang2022Finger}. Our results suggest that further consideration should include viscoelastic relaxation to achieve tactile sensing with fine motor control for object manipulation, especially in scenarios of stress relaxation when the fingers are holding the objects under a fixed position command to close the gripper.

    Creep is another phenomenon of viscoelasticity, which measures the time-dependent strain $\epsilon(t)$ under a constant force $\sigma_0$ applied on the soft material. The most common scenario is weight compensation while holding an object, which usually occurs on the network's secondary interaction face while holding an object. Using the experimental setup in Figure \ref{fig6}D, we place a cylindrical rod of a 15 mm radius at the center of the network's secondary interaction face. The soft network is tilted at $\gamma = 8^\circ$ to keep the secondary interaction face horizontal as the contact begins. By attaching different weights to the cylindrical rod, we tested its viscoelastic responses to small, medium, and large static forces of $F_y$ in 1.5, 3, and 5.9 N. Figure \ref{fig6}A (iii and iv) captures the creep effect with increased marker displacement along the $y$-axis. In contrast, the reaction force $F_y$ immediately reaches a stable state after placing the rod. Figure \ref{fig6}E shows the time-dependent creep compliance curves $C_{crp}(t)=\epsilon(t)/\sigma_0$. We model the creep phenomenon using a Kelvin model, composed of a spring of stiffness 1/$m_g$ in series with three Voigt elements as shown in Figure \ref{fig7}B. Each Voigt element consists of a Hookean spring of stiffness $1/m$ parallel to a Newtonian dashpot of viscosity $1/\varphi$, resulting in a characteristic time $\tau=m/\varphi$. The fitted Kelvin model is described by
    \begin{equation}
        C_{crp}(t)= m_g + \sum_{j=1}^3 m_j[1-\exp{(-t/\tau_j)}], \label{eq3}
    \end{equation}
    where t = 0 s is when external loading is completed, the glassy compliance $m_g$ = 0.97 mm/N, the three Voigt elements $m_1$, $m_2$, $m_3$ = 0.10, 0.11, 0.15 mm/N and the characteristic creep times $\tau_1$, $\tau_2$, $\tau_3$ = 3.1, 22.8, 206.2 s, respectively. The equilibrium compliance $C_{crp}(\infty)$ increased by a significant 37$\%$ compared to the initial compliance $C_{crp}(0)$. One can view creep as a reciprocal effect of relaxation. Both characterize the viscoelastic behavior of the network's molecular adaptation during static interaction. The experimental results agree well with the fact that the relaxation response is faster than creep \citep{Gutierrez-Lemini2013Engineering}.  
    
    We also conducted dynamic loading experiments using the setup in Figure \ref{fig6}F to investigate the Soft Polyhedral Network's nonlinear and viscoelastic behaviors when compressed by a flat probe at different depths, time intervals, and speeds. We recorded the fiducial marker's $x$-axis displacement in cyclic interactions, about 80$\%$ of the flat probe's pushing depth. At an average rate of 10 mm/s, loading and unloading cycles vary at depths between 2 and 20 mm, as shown in Figure \ref{fig6}G(i). The soft network shows non-negligible hysteresis in all experiments. We examined its recovery from deformation by applying consecutive loading and unloading cycles at 10 mm depth with various waiting times ranging from 0.5 to 20 s. As shown in Figure \ref{fig6}G(ii), right after the flat probe disengages with the soft network during unloading, we observe a minor residual strain decreasing rapidly as no further plastic deformation is observable after a waiting time of 5 s. For the two follow-up cycles with a waiting time of 0.5 s, their hysteresis loops almost overlapped even though a residual strain of 0.2 mm existed, demonstrating the soft network's robust performance against fatigue. To investigate the viscoelasticity of rate dependency, we probed the soft network at different rates ranging from the slowest 0.1 mm/s to the fastest 20 mm/s. Results in Figure \ref{fig6}G(iii) show that the stiffness increases to 25$\%$ as the loading rate increases, indicating the non-negligible viscoelastic effects in soft robotic interactions at different speeds.

\subsection{Visual Force Learning for Viscoelastic Proprioception} 
\label{sec:Method-VisualForceLearning}

    Sense of force and effort is another characteristic of proprioception by measuring or reproducing the absolute amount or the relative percentage of force applied. 
    \begin{figure*}[!ht]
        \includegraphics[width=\textwidth]{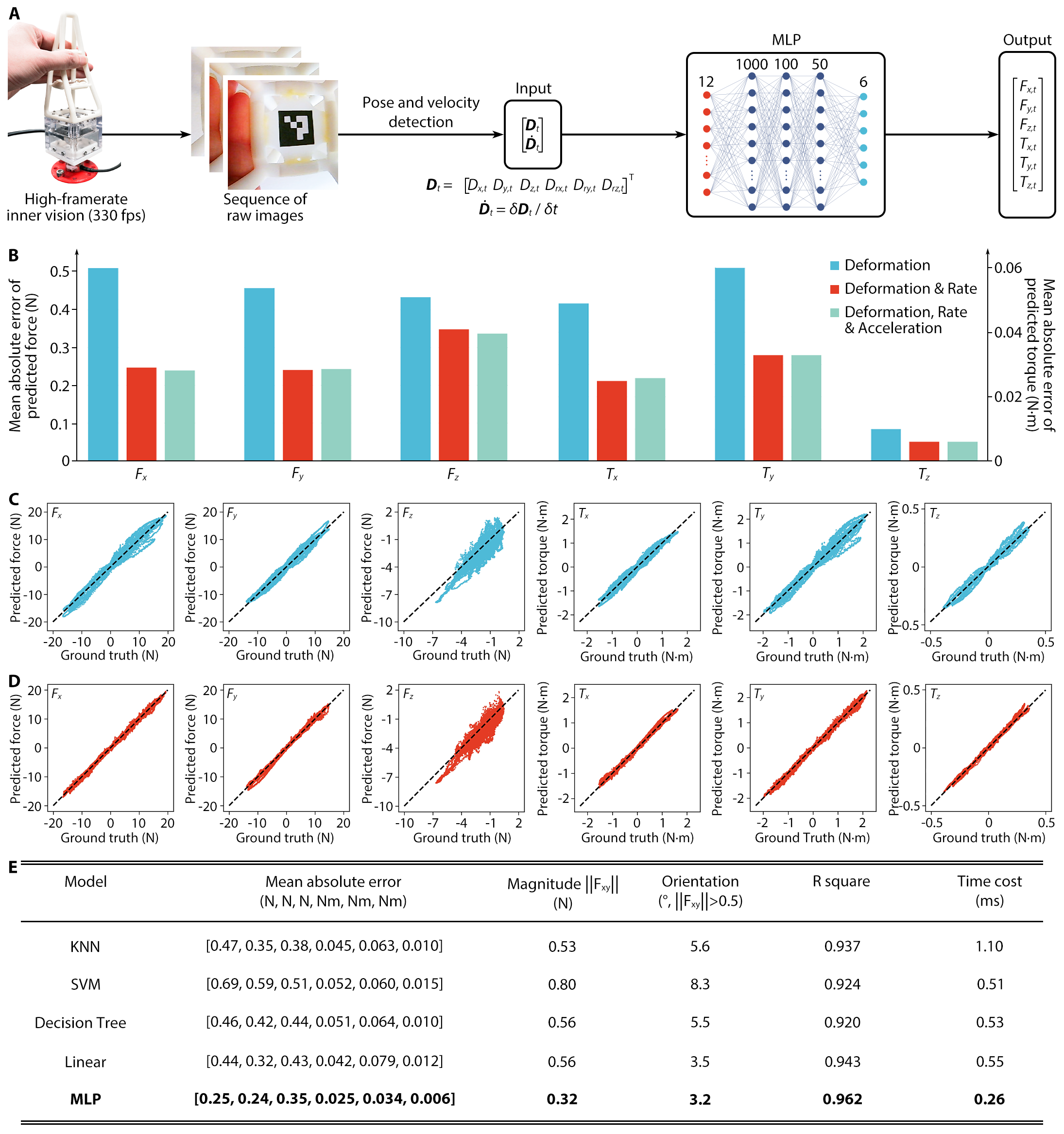}
        \caption{
        \textbf{Visual force learning for viscoelastic proprioception.} 
        (A) The proprioceptive learning pipeline, which starts by image processing for the marker's pose $\mathbf{D}_t$ relative to the initial state, then feeds pose $\mathbf{D}_t$ and its time derivative $\dot{\mathbf{D}}_t$ to an MLP model to infer 6D forces and torques. (B) The mean absolute errors of the predicted 6D forces and torques using input features including deformation only, deformation \& rate, and deformation, rate \& acceleration. (C) When trained with the deformation feature only, the forces and torques predicted by the MLP model exhibit observable hysteresis compared to the ground truth. (D) Similar to (C) but trained with deformation and rate as input features, where the hysteresis is largely eliminated. (E) Benchmark when using different models for learning, where the MLP outperformed the rest in all metrics.}
        \label{fig8}
    \end{figure*}
    Based on findings on viscoelasticity, we propose a visual force learning method to achieve viscoelastic proprioception by incorporating the Soft Polyhedral Network's kinetic motions. The overall framework is shown in Figure \ref{fig8}A, where the marker inside the network works like a physical encoder to convert passive, spatial deformations into a 6D pose vector $\mathbf{D}_t$, tracked by the miniature motion capture system inside the Soft Polyhedral Network. Then, we developed a decoder model using MLP to infer the corresponding 6D forces and torques ($F_x$, $F_y$, $F_z$, $T_x$, $T_y$, $T_z$) as the output. To reflect the speed of interaction during physical contact, we added a velocity term $\dot{\mathbf{D}}_t=\delta \mathbf{D}_t/\delta t$ to the model input by setting $\delta t$ = 15 ms (or five frames per interval at 330 fps) for more stable tracking. 
    
    The simplicity of design enabled us to collect 140,000 samples within 10 minutes by manually interacting with the Soft Polyhedral Network at different heights and speeds. We completed the data collection process within 10 minutes by manually interacting with the Soft Polyhedral Network at different heights ($H_2$ and $H_3$) and speeds (fast and slow). The collected dataset has 80,000 samples for training and 60,000 for testing, including the frame-by-frame raw images, recognized marker poses, 6D forces and torques from the ATI sensor as the true labels, and the corresponding timestamps. The dataset's maximum 6D forces and torques are 20 N, 20 N, 10 N, 2 Nm, 2 Nm, and 0.5 Nm, respectively. We normalized inputs and outputs within $[-1, 1]$ to balance the loss optimization in each dimension for more stable model predictions. The MLP consists of four hidden layers with 1,000, 100, 50, and 6 neurons, respectively (implemented in PyTorch), and were trained with a batch size of 32 using an Adam optimizer with a learning rate of 0.001 on mean squared error loss. We trained the models for 60 epochs and used the one that performed the best on the test dataset.
    
    We trained three models to verify the speed of interactions' contribution in minimizing the predictions' mean absolute errors (MAEs) for the 6D force and torque outputs in Figure \ref{fig8}B. When using deformation $\mathbf{D}_t$ as the only input, the model's mean absolute errors are 0.51 / 0.46 / 0.43 N ($F_x/F_y/F_z$) in forces and 0.049 / 0.062 / 0.01 Nm ($T_x/T_y/T_z$) in torques. However, after adding deformation rate $\dot{\mathbf{D}}_t$ to the input features, the prediction errors are reduced by almost half to 0.25 / 0.24 / 0.35 N in forces and 0.025 / 0.034 / 0.006 Nm in torques. The minor improvement in $F_z$ could be caused by the soft network's relatively less adaptiveness along the $z$-axis by design. Further adding deformation acceleration $\Ddot{\mathbf{D}}_t$ to the input features leads to a slight improvement in performance, suggesting that the ($\mathbf{D}_t, \dot{\mathbf{D}}_t$) inputs are sufficiently effective to achieve enhanced visual force learning for viscoelastic proprioception. 
    
    We further investigated the hysteresis error of the visual force learning model through comparison against the ground truth using ATI measurements. When using deformation $\mathbf{D}_t$ as the only input, we observed a hysteresis loop in all predicted force and torque components, as shown in Figure \ref{fig8}C. However, results in Figure \ref{fig8}D show that adding deformation rate $\dot{\mathbf{D}}_t$ to the input features eliminates the hysteresis effects substantially, where the prediction accuracies are consistently stable over different ranges of interactions. The MLP model is computationally efficient, with an average prediction time of 0.26 ms. The chosen camera's highest fps currently sets the upper bound of the sensing frequency at 330 Hz, which is still much higher than many existing vision-based tactile sensors from the research literature \citep{Sun2022Soft, Xu2021Compliant} or commercial products (such as the FT 300 from Robotiq) that run at 100 Hz or less. Higher force-sensing bandwidth is usually preferred for reactive control in real-time physical interactions. 

    We also benchmarked the MLP with four other models using ($\mathbf{D}_t, \dot{\mathbf{D}}_t$) as input, including KNN (K-Nearest-Neighbors), SVM (Support Vector Machine), Decision Tree, and Linear regression. We used the MAEs of forces and torques, the force magnitude accuracy $\|F_{xy}\|$ and force directional accuracy $\phi$ of MAEs in the $x$-$y$ plane, R-square, and computation time as the evaluation metrics. Results in Figure \ref{fig8}E suggest that the MLP model performs the best in all metrics. The force magnitude and directional accuracies are 0.32 N and 3.2 degrees, respectively. Since the force directional accuracy $\phi$ is extremely sensitive to disturbances in $F_x$ or $F_y$ for samples with small magnitudes, the calculation of directional MAEs only includes examples where force magnitude exceeds 0.5 N. We conducted an ablation study and found that adding kinetic features is crucial in eliminating the hysteresis effect even with linear regressor. At the same time, a more advanced model such as MLP further improves the overall performance. 

\section{Fine-Motor Skills in Object Handling}
\label{sec:Results}

    The overall framework for vision-based proprioceptive learning with the Soft Polyhedral Network proposed in this study consists of three major components, including the Sim2Real learning for kinesthesia adaption, viscoelastic modeling, and visual force learning for viscoelastic proprioception. This section applies the proposed proprioceptive learning with Soft Polyhedral Networks in fine-motor control for object manipulation tasks such as 1) sensitive and competitive grasping and 2) touch-based geometry reconstruction.

\subsection{Sensitive and Competitive Grasping against Rigid Grippers}
\label{sec:Results-PropLearningDemos-CompetitiveGrasping}

\begin{figure}[!p]
        \includegraphics[width=\textwidth]{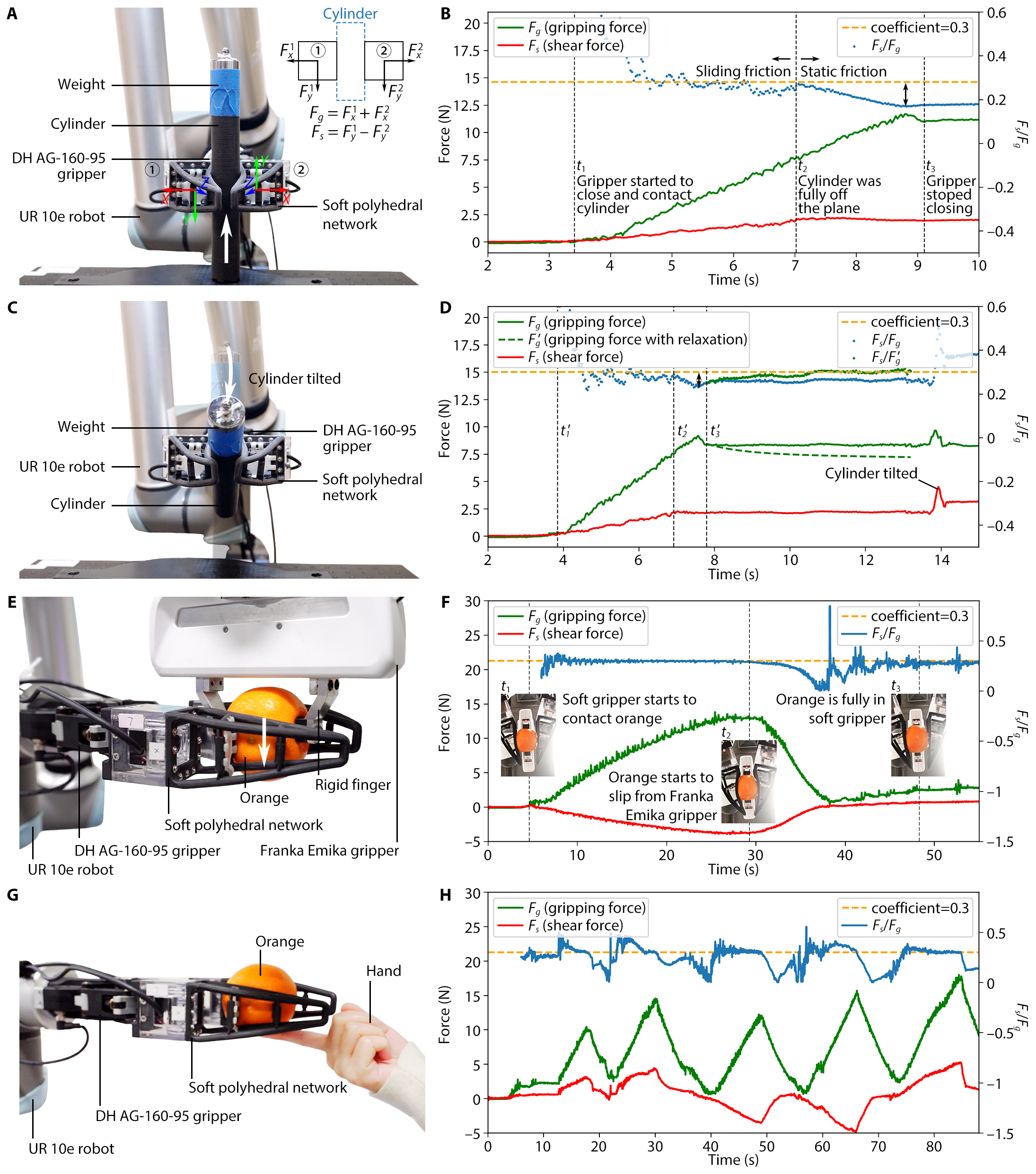}
        \caption{
        \textbf{Sensitive and robust grasping of the Soft Polyhedral Networks as robotic fingers.} 
        (A) Setup for a cylinder grasping task by replacing the gripper's original rigid fingertips with Soft Polyhedral Networks. (B) The measured gripping and shear forces, where the coefficient of sliding friction $\mu$ is estimated by $F_s/F_g$. The gripper produces enough friction against the cylinder's gravity to lift it off the table. (C)$\sim$(D) The same cylinder grasping task but with a gripping force approaching the frictional cone's boundary. The cylinder is lifted initially but tilts because the gripping force decreases due to viscoelastic relaxation and exceeds the friction cone. (E)$\sim$(F) Setup and results of an orange grasping task completed in a force control loop by maintaining the gripping and shear forces within a predefined friction cone, where the soft fingers successfully grabbed the orange from the rigid fingers. (G)$\sim$(H) Setup and results for an orange grasping task against disturbance from a human hand, where the soft fingers protect the orange from pushing attempts made by the human fingers.}
        \label{fig9}
    \end{figure}

    We demonstrate the superior performance of the Soft Polyhedral Networks as force-sensitive fingers for rigid grippers. Modern end-effectors, such as the two-finger gripper (Model AG-160-95 by DH-Robotics) shown in Figure \ref{fig1}D, usually come with removable, rigid fingertips that can be easily replaced with the proposed Soft Polyhedral Networks. We fabricated two new prototypes in black color to replace the original rigid fingers of a DH gripper, which was then installed on a collaborative robot (UR10e from Universal Robots) for manipulation tasks in Figure \ref{fig9}. The force/torque sensing model trained from the previous white soft network is directly transferable to the newly fabricated black ones, suggesting the soft networks' scalability with consistent performances in proprioceptive learning.

    In the first experiment, we demonstrate the soft network's capability in friction estimation during dynamic grasping by gradually closing the fingers while moving upward to pick up a 3D-printed cylinder. Figure \ref{fig9}A shows the two fingers' coordinate systems. The contact starts at $t_1$ with an increasing gripping force $F_g=(F^1_x+F^2_x)\cos\beta$ detected by the two Soft Polyhedral Networks with different superscripts (Figure \ref{fig9}B). Between $t_1$ and $t_2$, the cylinder did not leave the tabletop, and the soft fingers slid along the cylinder. We estimated the sliding friction's coefficient $\mu=F_s/F_g \approx 0.3$, where the estimated shear force $F_s=F^1_y-F^2_y$ equals the sliding friction. At $t_2$, $F_s$ exceeded the cylinder's gravity $G_{cyl}$, and the sliding friction became static with the cylinder lifted off the table while moving together with the gripper. At $t_3$, the gripper stopped closing, and the gripping force reached the maximum. In summary, the Soft Polyhedral Network's proprioceptive capability was sufficiently accurate to deal with dynamic loadings for friction estimation. However, when the soft gripper is holding the object for a more extended period in Figure \ref{fig9}D, the materials' stress relaxation causes the actual gripping force $F'_g$ to drop after $t_3$ according to $F_g'(t)=F_g(t_3) E_{rel}(t-t_3)/E_{rel}(0)$ where $E_{rel}$ is defined in Equation (\ref{eq2}) and $F_g$ is the MLP predicted force. As a result, the ratio $F_s/F'_g$ increased and approached the friction cone's boundary, causing the cylinder to tilt with weight imbalance at $t \approx$ 14 s in Figure \ref{fig9}D (See supplementary material Movie S2). 
    
    We further demonstrate the soft network's capability in competitive grasping against rigid grippers and a human finger. The experiment in Figure \ref{fig9}E begins with a Franka Emika holding an orange with its rigid gripper. The gripper with the Soft Polyhedral Networks on a UR10e closes, intending to pull the orange away by moving downward. After contacting the orange at $t_1$, the gripper needs to actively adjust its gripping width based on sensory feedback from the Soft Polyhedral Networks in a force control loop to maintain the $F_s/F_g$ ratio within a predefined friction cone (Figure \ref{fig9}F). Both gripping and shear forces increased simultaneously until the sum of $F_s$ (4 N) and orange's gravity $G_{org}$ exceeded the friction on Franka's gripper at $t_2$, indicating the moment when the orange started to slip from the rigid gripper to the soft fingers. Then, the shear force decreased and changed direction to counteract the orange's weight while the orange was fully secured within the soft fingers at $t_3$ (See supplementary material Movie S3). We also conduct a follow-up experiment in Figure \ref{fig9}G by manually pushing the orange out of the soft fingers. Results in Figure \ref{fig9}H demonstrate the Soft Polyhedral Networks' dynamic capability in proprioceptive learning to retain the orange from four attempts by the human finger with a maximum gripping force of 18 N.

    \begin{figure}[!p]
        \includegraphics[width=\textwidth]{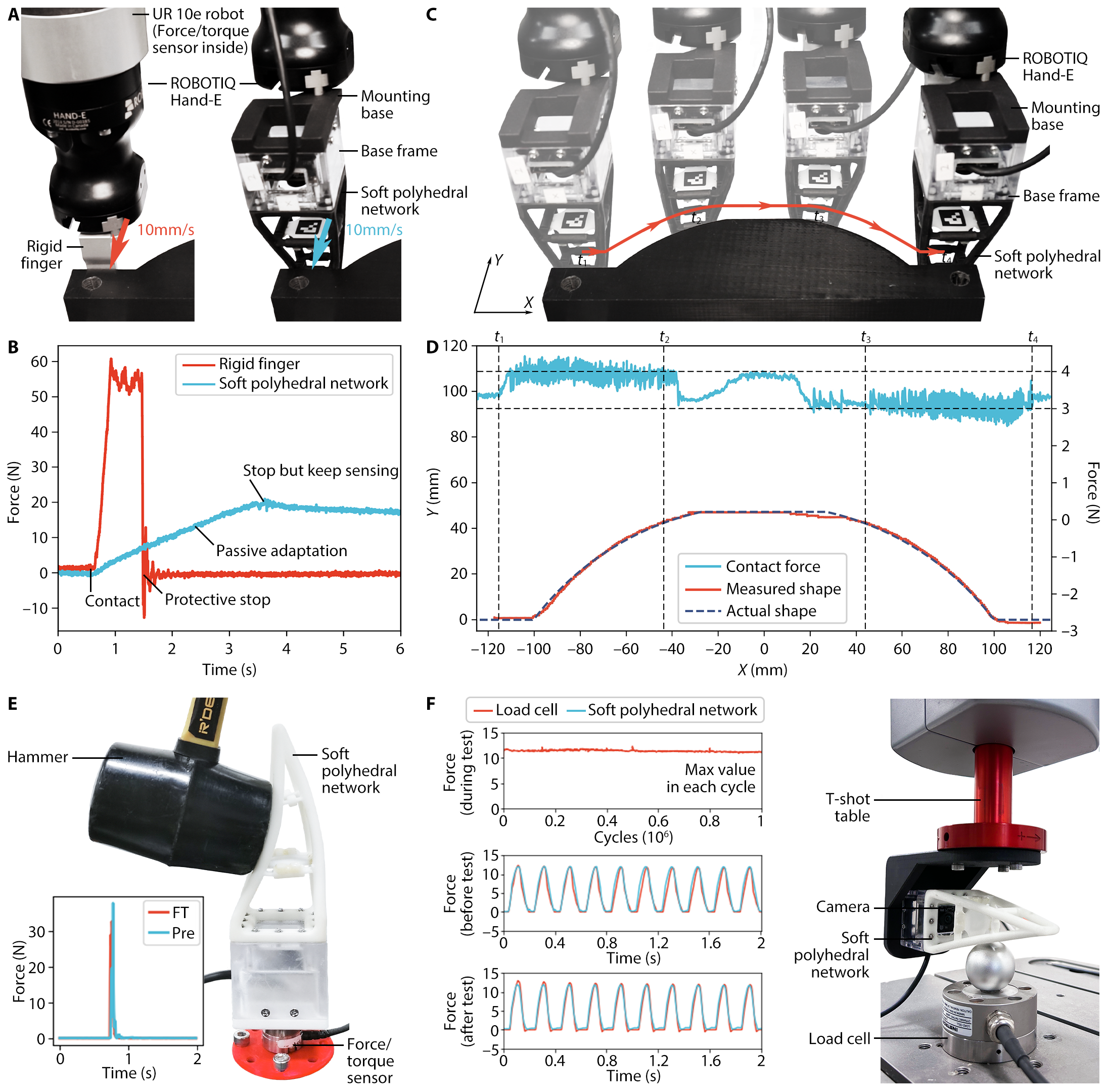}
        \caption{
        \textbf{Tactile reconstruction and impact absorption of the Soft Polyhedral Network.}
        (A) Setup for impact absorption comparison between the original rigid finger and a Soft Polyhedral Network as the finger when the fingers hit a rigid obstacle at 10 mm/s. (B) The rigid collision generates a 60 N impact force that instantly triggers a protective stop in the robot controller, whereas the soft collision only generates 20 N within 3 s. (C) Setup for touch-base geometry reconstruction using the soft network as the finger by sliding along the arch contour of a 3D-printed object by maintaining the contact force $F_{xy}$ between 3 to 4 N. (D) Experiment results of the reconstruction by touch indicating the contact force in light blue lines, and the gripper trajectory as the reproduced geometry in red lines. (E) When hit by a hammer, the soft network remains functional while accurately measuring the impact force. (F) Setup and results for fatigue test. The soft network moved downward at a frequency of 5 Hz for one million cycles and maintained stable mechanical properties and proprioceptive prediction ability.}
        \label{fig10}
    \end{figure}

\subsection{Impact Absorption and Touch-based Geometry Reconstruction}
\label{sec:Results-PropLearningDemos-ImpactAbsorption}

    The Soft Polyhedral Network's proprioceptive capability can also sense and absorb impacts during a collision, supporting continuous task completion without interruption, representing a practical demand from modern production lines with robots (See supplementary material Movie S4). For a collaborative robot (UR10e from Universal Robots) with rigid fingers on its gripper (Hand-E from Robotiq) in Figure \ref{fig10}A, 

    a collision of 10 mm/s incurs an impact force up to 60 N for nearly 1 s, measured by the F/T sensor inside the flange until the protective stop is triggered in the controller in Figure \ref{fig10}B. However, after replacing the rigid fingertip with the soft network, the measured impact was much reduced to one-third the amount (20 N) within three times the duration, resulting in a combined 9X improvement in safety factor. The soft network's passive deformation effectively absorbed the impact without causing an emergency stop, enabling the system to continue with predefined tasks, such as the tactile reconstruction task shown in Figure \ref{fig10}C. In this experiment, we implemented a force control strategy by keeping the contact force between the soft network and the arch shape workpiece between 3 and 4 N in the $x$-$y$ plane, letting the soft network slide along the target object's contour for geometry reconstruction. Figure \ref{fig10}D shows that the recorded trajectory skillfully reconstructs the target object's geometry. 

    We also tested the robustness and durability of the Soft Polyhedral Network's proprioceptive prediction under sudden or repetitive impacts. In Figure \ref{fig10}E, we struck the primary interaction surface using a hammer while mounting the network on top of an F/T sensor (Nano25 from ATI) on a table. The F/T sensor detected a sudden impact up to 35 N within 34 ms, and the predicted result matched the sensor readings with a robust performance. In Figure \ref{fig10}F, we performed a fatigue test for the Soft Polyhedral Network on the Instron$^\circledR$ ElectroPuls$^\circledR$ E3000. The soft network was mounted on the T-shot table and moved cyclically downward at a frequency of 5 Hz, with a maximum displacement of 10 mm and a resultant force of 13 N. The soft network contacted the rigid ball fixed on the load cell, causing its deformation. During the one-million-cycle loading process, the contact forces recorded by the load cell remained between 11.0 and 11.8 N, demonstrating that the network maintained stable mechanical properties (See supplementary material Movie S5). In addition, the soft network's predicted forces agree well with the forces recorded by the load cell before and after one million loading cycles, proving that the fingers had robust and durable proprioceptive prediction ability.

\section{Conclusion, Limitation, and Future Work}
\label{sec:Final}

    The medical literature often regards proprioception as the sixth sense that tells us what the body itself is doing. It involves the sense of position and movement and the sense of force and effort through our musculoskeletal system, a skill essential for the robot to acquire for intelligent interaction with the physical world. During the moment of touch, the sensing receptors under our skin detect the object's physical properties through a mixture of modalities and simultaneously react by adjusting the muscle contraction with skeletal movement to facilitate a natural interaction from within. In comparison, classical design methods through rigid-body mechanics excel in accuracy and speed, and emerging solutions in soft robotics support an overconstrained interaction through under-actuated designs that are passively adaptive to the unstructured environment with mechanical intelligence. In this study, we proposed the design of a class of Soft Polyhedral Networks capable of whole-body compliance adaptive to external interactions with an embedded vision-based motion capture system inside. We achieved adaptive kinesthesia using Sim2Real learning from FEM data to reproduce the soft network's whole-body position and movement during passive adaptation in real-time. We also proposed a visual force learning method for viscoelastic proprioception by adding velocity terms to the positional input features to infer more accurate senses of force and effort using neural networks. The prototypes presented in this study use only one off-the-shelf camera board with two 3D-printed components while being functionally compliant in 3D with sensory feedback in 6D, much cheaper than commercial 6-axis force-torque sensors to facilitate mass adoption. Within a compact form factor and a wide range of design variations, one can easily customize the Soft Polyhedral Network to suit the changing needs for force-controllable interactions in modern robotics with transferable, robust, real-time proprioceptive learning.

    The polyhedron-inspired network design proposed in this work is a versatile method to introduce customizable spatial compliance to physical interactions in robotics with ample design space for further optimization. We selected a particular design in this study with enhanced performance for grasping, featuring a primary interaction face with a larger contact area for adaptive grasping and a secondary one to facilitate spatial compliance. One can degenerate the design to 2D compliance by changing the soft network into a multi-layered structure, resulting in a design like Festo's Fin-ray finger. Such degenerated 2D structure is limited to purely planar adaptation only with an obstructed interior view, challenging for vision integration in applications \citep{Xu2021Compliant}. The polyhedron-inspired geometry greatly enhanced the design variations while enabling spatial adaptation to preserve the proprioceptive learning capability with sense. We can also add layers of friction-resistive material \citep{Li2022Optimization, Li2020Effect} or introduce bio-inspired texture \citep{Zhang2020Micro} directly on the primary interaction face to further enhance the Soft Polyhedral Network's performances in grasping. One can also modify the design to make the beams hollow, allowing fluidic actuation to adjust the stiffness distribution of the whole network to maximize the power of soft robotics, but at the cost of added complexities in pressurized fluidic power source and system design.

    The Soft Polyhedral Network is simple, accurate, and robust in producing stable mechanical properties and proprioceptive predictions  even after one million compression cycles, accommodating physical interactions for robotic manipulation in tasks such as sensitive and competitive grasping and touch-based geometry reconstruction. Considering the low cost of the material and parts and the relatively large load of each push presented in the experiments, our design is remarkably durable. It is worth noting that only the soft network part needs replacement during maintenance, while the camera base and learning algorithms can be reused. The off-the-shelf camera board currently limits the size, frequency, and processing power of the Soft Polyhedral Network. With added cost in custom development, we can further upgrade the camera with a higher framerate in a smaller size, use battery-powered onboard processing for edge computing and wireless communication, and introduce active lighting with LEDs for a more stable capture of image features. We also point out that the pyramid design used in this study is limited in adaptation along the $z$-axis, mainly chosen with an enhanced adaptation in the $x$-$y$ plane for grasping, resulting in a less accurate force estimation along the $z$-axis. This issue can be addressed using different network designs for passive adaptation in desirable axes.

    The simplicity of the Soft Polyhedral Network design strengthens the integration of visual features to support learning-based capabilities in a more challenging environment. For example, with simple waterproofing of the base mount and the camera inside, the system can be directly used underwater while maintaining proprioception. One can obliterate the marker by processing full images of the soft network deformations with advanced neural networks, such as variational auto-encoders, for proprioceptive learning but at the cost of explainability, transferability, and accuracy. Recent advances in generative models could also be a promising solution to automatically remove the network in the image with generated pixels of the physical world so that the camera can be alternatively used as a vision sensor for object detection. Investigations into the viscoelastic behaviors of the Soft Polyhedral Networks demonstrate the importance of including kinetic features while integrating machine learning with soft robots. We can use a single high-framerate vision sensor to capture the soft network's dynamic physical interaction process with a rich collection of visual features to support a learning-based approach. With the vision-based solution and learning algorithms, if relaxing the need for functional passive adaptation, one can use almost any deformable, hollow structure on top of the camera to achieve proprioceptive learning presented in this work, as long as a reasonable volume inside is within the camera's viewing angles during physical interactions.

\section*{Acknowledgements}

    This work is partly funded by the National Natural Science Foundation of China under Grant 62206119, the Science, Technology, and Innovation Commission of Shenzhen Municipality under Grants ZDSYS20220527171403009 and JCYJ20220818100417038, and Guangdong Provincial Key Laboratory of Human-Augmentation and Rehabilitation Robotics in Universities.

\section*{Supporting Videos}
\label{sec:SM-Videos}

    \begin{itemize}
        \item Movie S1. Sim2Real proprioception for adaptive kinesthesia.
        \item Movie S2. Viscoelastic sensitive grasping for friction estimation.
        \item Movie S3. Competitive grasping for an orange.
        \item Movie S4. Impact absorption and tactile reconstruction using visual force learning.
        \item Movie S5. Million cycle fatigue test.
    \end{itemize}

\bibliographystyle{unsrtnat}
\bibliography{references}

\end{document}